\documentclass[11pt, letterpaper]{article}
\usepackage[utf8]{inputenc}
\usepackage[margin=1.5cm]{geometry}
\usepackage{titlesec}
\usepackage{tabu}
\usepackage{enumitem}
\usepackage{amssymb}
\usepackage{xcolor}
\newlist{selectlist}{itemize}{2}
\setlist[selectlist]{label=$\square$,leftmargin=*,noitemsep,topsep=0pt}

\usepackage{lmodern}

\usepackage{hyperref}
\hypersetup{
    colorlinks=true,
    linkcolor=blue,
    filecolor=magenta,      
    urlcolor=blue,
}
\newcommand{\contentBasePath}[0]{content/}
\newcommand{\visBasePath}[0]{\contentBasePath vis/}
\usepackage[utf8]{inputenc} % Input encoding utf8
\usepackage[english]{babel}
\usepackage{amsmath}
\usepackage{amssymb}
\usepackage{graphicx} % Graphics http://ctan.org/pkg/graphicx
\usepackage{xspace} % for xspaces in new commands
\usepackage{todonotes}
\usepackage{subfig} % For using subfigures and alternative to subcaption if there are template issues
\usepackage[normalem]{ulem}
\usepackage{yfonts}
\usepackage[backend=biber, style=numeric-comp, sorting=nyt, maxbibnames=10, natbib=true]{biblatex}
\usepackage{wrapfig}
\usepackage[export]{adjustbox}
\usepackage{hyperref}
\hypersetup{breaklinks=true}

\definecolor{deepblue}{rgb}{0,0,0.5}
\definecolor{deepred}{rgb}{0.6,0,0}
\definecolor{deepgreen}{rgb}{0,0.5,0}

\DeclareFixedFont{\ttb}{T1}{txtt}{bx}{n}{8} % for bold
\DeclareFixedFont{\ttm}{T1}{txtt}{m}{n}{8}  % for normal

\usepackage{listings}
\newcommand\pythonstyle{\lstset{
language=Python,
basicstyle=\ttm,
breaklines=true,
commentstyle=\small\textit,
morekeywords={self},              % Add keywords here
keywordstyle=\ttb\color{deepblue},
emph={MyClass,__init__},          % Custom highlighting
emphstyle=\ttb\color{deepred},    % Custom highlighting style
stringstyle=\color{deepgreen},
frame=tb,                         % Any extra options here
showstringspaces=false
}}
\lstnewenvironment{python}[1][]
{
\pythonstyle
\lstset{#1}
}
{}
\newcommand\pythoninline[1]{{\pythonstyle\lstinline!#1!}}

\newboolean{vc_is_included}
\InputIfFileExists{vc.tex}{
	\setboolean{vc_is_included}{true}
}{
	\setboolean{vc_is_included}{false}

}

\newboolean{article_extended}
\ifthenelse{\isundefined{\makeExtended}}{% Configure \makeExtended with command from Makefile
	\setboolean{article_extended}{false}}{
	\setboolean{article_extended}{true}}

\newcommand{\articleAuthor}[0]{Julian Stier, Michael Granitzer}

\titleformat{\section}[block]{\hspace{1em}\bfseries}{\thesection.}{0.5em}{} 
\titleformat{\subsection}[block]{\hspace{1em}}{\thesubsection}{0.5em}{}

\begin{document}
\noindent
\textbf{\textit{deepstruct -- linking deep learning and graph theory}}
\vskip0.5cm
\noindent
\textbf{\textit{\articleAuthor}\\
	Chair of Data Science, University of Passau\\
	\texttt{julian.stier@uni-passau.de}
}\\

\noindent
\textbf{Abstract}\\
\texttt{deepstruct} connects deep learning models and graph theory such that different graph structures can be imposed on neural networks or graph structures can be extracted from trained neural network models.
For this, \texttt{deepstruct} provides deep neural network models with different restrictions which can be created based on an initial graph.
Further, tools to extract graph structures from trained models are available.
This step of extracting graphs can be computationally expensive even for models of just a few dozen thousand parameters and poses a challenging problem.

\texttt{deepstruct} supports research in pruning, neural architecture search, automated network design and structure analysis of neural networks.

\vskip0.5cm

\noindent
\section{Analysing Graph Structures of Neural Networks}
% A short description of the high-level functionality and purpose of the software for a diverse, non-specialist audience
Neural network architectures become larger and larger with thousands to even billions of parameters.
A lot of effort therefore is put into either extracting high-level design information out of automatically found architectures or to impose architectural design on models to achieve high accuracy, high performance, low energy or low memory consumption.
From a formal perspective, an architectural design can be understood as directed acyclic graphs with or without labels and certain restrictions.
With \texttt{deepstruct}, deep learning models and graphs are combined on top of the famous \texttt{pytorch} \cite{paszke2019pytorch} and \texttt{networkx} \cite{hagberg2008exploring} libraries such that they can be transformed into each other.

%mg model families below is not clear for me. Maybe refer back to the formal perspective of the architectural design and then reuse it below/
\texttt{deepstruct} is a python package providing model architectures of neural networks, building blocks to create own sparse networks and a transformation process from trained neural networks into graphs.
Deep neural network models can be created based on directed graphs.
The package aims to provide a complete round-trip for transforming graphs into sparse neural networks and back.
In terms of neuroevolution, which aims to build neural networks in an automated and evolutionary way, this can be seen as part of the transformation from genotypical into phenotypical encoding.
The discrete graph space then forms the genotypical encoding which is transformed into a deep neural network architecture and trained on data.

\section{Using \texttt{deepstruct}}
The package provides several models which have built-in binary masks to impose a graph structure across layers.
These masks can be configured prior to training or be re-computed e.g. through pruning.
Multiple masks over a layered model constitute a directed acyclic graph.
A graph can either be used to initialize or modify the masks or it can be extracted from trained models.

\texttt{deepstruct} serves multiple purposes in context of bridging a gap between graphs and sparse neural network models.
An example is to \textit{prune} an initially fully connected multi-layered neural network to find lottery tickets \cite{frankle2018lottery}.
Lottery tickets are a combination of an initial weight distribution and binary masks obtained through iterative magnitude-based pruning.
These binary masks can be easily extracted from \texttt{MaskedLayer}s.
Another practical example is to \textit{search} for an optimal graph from a pre-defined graph family (a design space) which is trained on a dataset when transformed into a deep neural network.

Steps to undertake from a graph to a deep neural network:
1) Choose a graph space, e.g. by sampling random Watts-Strogatz\cite{watts1998collective} graphs.
2) Choose an appropriate model, e.g. directed acyclic deep networks with skip-layer connections and with graph vertices being translated to neurons: \texttt{deepstruct.sparse.MaskedDeepDAN}.
The model is chosen based on the restrictions imposed on the model.
Purely feed-forward models without skip-layer connections should choose \texttt{MaskedDeepFFN} instead of \texttt{MaskedDeepDAN}.
3) Construct the model given its structure, e.g. \texttt{m = deepstruct.sparse.MaskedDeepDAN(784, 10, structure)}.

Steps to undertake from a deep neural network to its graph structure:
1) Start with a \texttt{pytorch} model \texttt{m} e.g. built with \texttt{deepstruct}.
2) Decide on which granularity level neural connections and neurons should be transformed into graph edges or vertices.
3) Initialize a graph transformation with a random input vector which is required to pass it once through the forward-inference mode of the model: \texttt{functor} = \texttt{GraphTransform(}\texttt{torch.randn((1, 28, 28))}.
4) Obtain the \texttt{networkx} graph structure via \texttt{graph = functor.transform(m)}.

\begin{python}
import networkx as nx
import deepstruct.sparse

# Use networkx to generate a random graph based on the Watts-Strogatz model
random_graph = nx.newman_watts_strogatz_graph(100, 4, 0.5)
structure = deepstruct.graph.CachedLayeredGraph()
structure.add_edges_from(random_graph.edges)
structure.add_nodes_from(random_graph.nodes)

# Build an MNIST-images classifier with 784 input and 10 output neurons and the given structure
model = deepstruct.sparse.MaskedDeepDAN(784, 10, structure)
model.apply_mask()  # Apply the mask on the weights (hard, not undoable)
model.recompute_mask()  # Use weight magnitude to recompute the mask from the network

# Get the structure -- a networkx graph -- based on the masks in the pruned model
functor = GraphTransform(torch.randn((1, 784))
pruned_structure = functor.transform(model)
new_model = deepstruct.sparse.MaskedDeepDAN(784, 10, pruned_structure)
\end{python}

\paragraph{Pruning}
Pruning \cite{stier2021experiments} refers to a set of techniques removing structural elements of deep neural networks such as removing connections, neurons or sets of neurons.
The result after a pruning operation is a modified structure which can be understood as reducing the parametric search space to a subset.
\texttt{deepstruct} supports pruning by providing binary masks over weight matrices and tools to modify them.
Several pruning strategies can be easily implemented by modifying these masks.

\paragraph{Neural Architecture Search}
Neural Architecture Search (NAS) \cite{wistuba2019survey} refers to finding high-level designs of deep neural networks, often in form of graphs.
While pruning is a top-down approach to such a search or is often motivated by compression, NAS is more concerned by the design space and how to effectively navigate through it both in constructive, destructive or hybrid manner.
\texttt{deepstruct} provides several models with very large search spaces which can be further restricted by constraints on used graphs.
These models can be automatically built based on a given graph and the transformation processes of \texttt{deepstruct} between modules and graphs can be easily integrated into neural architecture search pipelines.

\begin{figure}[tb]
	\centering
	\includegraphics[width=0.9\textwidth]{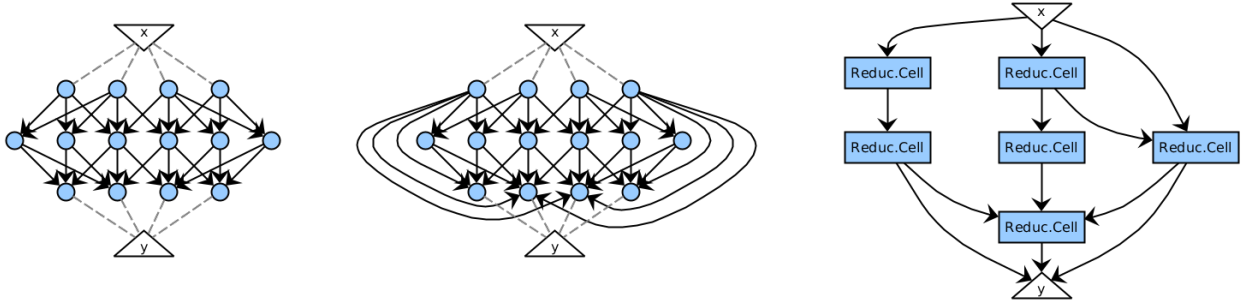}
    \caption{
		Sketches of a feed-forward architecture (left), a directed acyclic network with skip-layer connections (center) and a directed acyclic network with high-level operations (right).
    }
	\label{fig:arches-sketch}
\end{figure}

\paragraph{Deep Sparse Feed-Forward Networks}
Deep feed-forward networks are highly effective models capable of universal approximation with growing size.
In cases in which no skip-layer connections are required deep sparse feed-forward networks can be implemented as successive layers of matrix multiplications and sparsity constraints can be controlled through binary masks.
This has memory improvements over deep directed acyclic networks.
The following model constitutes a deep feed-forward network with six layers of decreasing size:
\begin{python}
model_ffn = deepstruct.sparse.MaskedDeepFFN(784, 10, [1000, 800, 500, 200, 50, 20])
\end{python}
This model is especially suitable for pruning analysis, e.g. to recompute the masks in all layers based on a weight-magnitude threshold $\theta=0.1$:
\begin{python}
model_ffn.recompute_mask(theta=0.1)  # Re-compute the mask based on a magnitude-based pruning strategy
for layer in deepstruct.sparse.maskable_layers(model):  # Iterate through all model layers with masks
    print(torch.sum(layer.mask)/float(torch.numel(layer.mask)))  # e.g. tensor(0.7204) -- density of a layer
\end{python}

\paragraph{Deep Directed Acyclic Networks}
To account for skip-layer connections the model requires up to $\frac{l^2-3l+2}{2}$ more matrices in the number of layers $l$ but changes from the feed-forward model of the form $y^l = \sigma(W_ly^{l-1}+B_l)$ to one with skip-layer connections of the form $y^l = \sigma(\sum^{l-1}_{j=0} W_{j\mapsto l}y^j+B_l)$ with $y^0 = x, l > 0$.
Any directed acyclic graph can be layered and brought into this form, allowing for varying layer sizes and skip-layer connections over multiple layers:
\begin{python}
deepstruct.sparse.MaskedDeepDAN(784, 10, structure)  # MNIST model with random generated structure
\end{python}
This model is particularly useful to investigate on graphs transformed into networks on the weight-level where each vertex of the graph corresponds to a neuron in the model.

\paragraph{Deep Cell-Based Networks}
Graphs of huge models on the weight level are difficult to analyse as they contain millions of vertices.
Such huge graphs can also consume huge amounts of memory and many graph operations grow non-linearly in the number of vertices.
Most neural architecture search strategies explore design spaces in which vertices of directed acylic graphs consitute high-level operations such as convolutions with kernels and weight sharing or poolings.
This decreases the graph drastically in size to only few, dozens or hundreds of vertices.
A single vertex then represents e.g. a mixture of operations we shorten as \textit{ReductionCell} and contains a convolution with a $5\times 5$ sized kernel, a ReLU activation function and a batch normalization layer.
A deep network with a given structure and high-level complex cell operations per vertex can then be constructed with a manually configurable constructor function:
\begin{python}
def my_cell_constructor(is_input, is_output, in_degree, out_degree, layer, input_channel_size):
    return ReductionCell(input_channel_size, 1) if is_input else ReductionCell(in_degree, 1)
deepstruct.sparse.DeepCellDAN(num_classes=10, input_channel_size=3, my_cell_constructor, structure)
\end{python}

\paragraph{Extracting Graph Structures From Deep Networks}
The opposite way around of extracting a graph structure from a deep neural networks requires some configuration for the general case of a \texttt{pytorch} module.
First, to support the automatic extraction, an initial input vector of fixed shape is required for a single feed-forward inference step.
During this step, \texttt{deepstruct} follows the actual computational structure of the model and prepares it to transform its modules into elements of the resulting graph.
Second, it needs to be defined on which granularity level the deep neural network should be translated back into a graph.
In the current default case of translating model weights into connections and neurons into vertices common state-of-the-art architectures result in huge graphs with the number of edges being in the number of the models' parameters -- millions to billions of vertices and edges.
This is a very memory-intensive task and also prevents many graph algorithms which are not operating in $\mathcal{O}(n)$ to take very long.
One alternative way is to translate higher-level elements such as computational modules or layers into vertices which might result in relatively small graphs with few information about the represented architecture family.
Another alternative is to conduct a graph compression (or simplification) routine after the low-level transformation with the goal to retain desired graph properties.
\begin{python}
shape_input = (3, 224, 224)
model = deepstruct.sparse.MaskedDeepDAN(784, 10, structure)  # Take any pre-trained deep learning model
functor = GraphTransform(torch.randn((1,) + shape_input))
graph = functor.transform(model)  # Auto-extract its graph structure
# Number of nodes, Number of edges, Overall graph density
print(len(graph.nodes), len(graph.edges), nx.density(graph))
\end{python}

\section{Related Work \& Applications}
Up to today we are not aware of a directly comparable framework which concentrates on providing a round-trip between graph representations and neural network architectures.
We assume the reason for this to be a young research field of Neural Architecture Search in which design spaces are automatically explored.
Further, transforming a deep learning model to a graph always means a loss of information as weight distributions within neural networks are reduced to their pure connectivity -- something which is assumably only favorable for exactly that purpose of working on neural architectures in this reduced space.

The Open Neural Network Exchange ONNX \cite{bai2019onnx} provides an exchange format for deep learning models and can be used to visualize or extract information from the underlying computational graph and the purpose can be considered more general and with a different focus.
TensorFlow Grappler \cite{larsen2019tensorflow} is the built-in optimizer for computational graphs of TensorFlow which applies transformations on the graph representation to gain speed-up, comparable to optimizations in the LLVM compiler infrastructure and the purpose can also be considered very differently but shows an application in a partially shared theoretical research area.

Applications of extracting graphs from a deep neural network comprise the visualization, interpretation, modification and meta-learning of depp neural network structures.

\section{Publications and Impact}
\texttt{deepstruct} fosters empirical research on the understanding of structures of deep neural networks.
Such research is conducted across multiple fields such as pruning, neural architecture search \cite{wistuba2019survey}, neuroevolution and theory on hidden stuctures of deep neural networks.
In particular, the package has been used in experiments of \cite{stier2019structural} and \cite{stier2021experiments} for correlation analyses between deep neural network performances and structural properties following the hypothesis that the deep hidden structure has an influence on the neural network performance or other properties such as robustness, memory consumption or hardware applicability.

Daily practice of empirical deep learning research is supported by shifting focus away from engineering \texttt{pytorch} modules towards defining structural design spaces for neural networks and investigating on methods to automatically navigate through such spaces.
This is achieved via automated model generation based on graphs and automated graph extraction from deep learning models.
In further developments, we aim to find solutions for high-memory usage in context of extracting graphs from large state of the art models with millions to billions of parameters and aim to provide more practical examples for graph visualizations and automated neural architecture design pipelines.

\paragraph{Acknowledgment}
Muhammad Yasir Mushtaq and Harshil Darji have contributed code to this library and Mehdi Ben Amor, Arsal Munir and Junaid Fahad contributed to verifications and usages during their master thesis research.

\vskip0.3cm

\defbibheading{bibintoc}[\bibname]{%
  \phantomsection
  \manualmark
  \markboth{\spacedlowsmallcaps{#1}}{\spacedlowsmallcaps{#1}}%
  \addtocontents{toc}{\protect\vspace{\beforebibskip}}%
  \addcontentsline{toc}{chapter}{\tocEntry{#1}}%
  \chapter*{#1}%
}
%\nocite{*} % cite all
\printbibliography

@article{watts1998collective,
  title={Collective dynamics of ‘small-world’networks},
  author={Watts, Duncan J and Strogatz, Steven H},
  journal={nature},
  volume={393},
  number={6684},
  pages={440},
  year={1998},
  publisher={Nature Publishing Group}
}

@techreport{hagberg2008exploring,
  title={Exploring network structure, dynamics, and function using NetworkX},
  author={Hagberg, Aric and Swart, Pieter and Schult, Daniel},
  year={2008},
  institution={Los Alamos National Lab.(LANL), Los Alamos, NM (United States)}
}

@article{frankle2018lottery,
  title={The lottery ticket hypothesis: Finding sparse, trainable neural networks},
  author={Frankle, Jonathan and Carbin, Michael},
  journal={arXiv preprint arXiv:1803.03635},
  year={2018}
}

@article{paszke2019pytorch,
  title={Pytorch: An imperative style, high-performance deep learning library},
  author={Paszke, Adam and Gross, Sam and Massa, Francisco and Lerer, Adam and Bradbury, James and Chanan, Gregory and Killeen, Trevor and Lin, Zeming and Gimelshein, Natalia and Antiga, Luca and others},
  journal={Advances in neural information processing systems},
  volume={32},
  pages={8026--8037},
  year={2019}
}

@article{wistuba2019survey,
  title={A survey on neural architecture search},
  author={Wistuba, Martin and Rawat, Ambrish and Pedapati, Tejaswini},
  journal={arXiv preprint arXiv:1905.01392},
  year={2019}
}

@article{stier2019structural,
  title={Structural Analysis of Sparse Neural Networks},
  author={Stier, Julian and Granitzer, Michael},
  journal={Procedia Computer Science},
  volume={159},
  pages={107--116},
  year={2019},
  publisher={Elsevier}
}

@misc{larsen2019tensorflow,
  title={TensorFlow Graph Optimizations},
  author={Rasmus Munk Larsen and Tatiana Shpeisman},
  year={2019}
}

@article{bai2019onnx,
  title={Onnx: Open neural network exchange},
  author={Bai, Junjie and Lu, Fang and Zhang, Ke and others},
  journal={GitHub repository},
  year={2019}
}

@article{stier2021experiments,
  author="Stier, Julian and Darji, Harshil and Granitzer, Michael",
  title="Experiments on Properties of Hidden Structures of Sparse Neural Networks",
  journal={arXiv preprint arXiv:2107.12917},
  year={2021}
}

\end{document}